\pdfoutput=1

\documentclass[11pt]{article}

\usepackage{ACL2023}

\usepackage{times}
\usepackage{latexsym}

\usepackage[T1]{fontenc}

\usepackage[utf8]{inputenc}

\usepackage{microtype}

\usepackage{inconsolata}
\usepackage{amsmath}
\usepackage{amssymb}
\usepackage{graphicx}

\title{Language Models Get a Gender Makeover: \\ Mitigating Gender Bias with Few-Shot Data Interventions}

\author{Himanshu Thakur \: Atishay Jain\thanks{\;\;Equal Contribution} \: Praneetha Vaddamanu\footnotemark[1] \\ \textbf{Paul Pu Liang} \: \textbf{Louis-Philippe Morency} \\ Carnegie Mellon University \\ {\texttt{\{\href{mailto:hthakur@andrew.cmu.edu}{hthakur},\href{mailto:atishayj@andrew.cmu.edu}{atishayj},\href{mailto:pvaddama@andrew.cmu.edu}{pvaddama},\href{mailto:pliang@andrew.cmu.edu}{pliang},\href{mailto:morency@andrew.cmu.edu}{morency}\}@andrew.cmu.edu}}}

\begin{document}
\maketitle

\begin{abstract}

\emph{Caution: this paper contains potentially offensive or upsetting model outputs.}

Societal biases present in pre-trained large language models are a critical issue as these models have been shown to propagate biases in countless downstream applications, rendering them unfair towards specific groups of people. Since large-scale retraining of these models from scratch is both time and compute-expensive, a variety of approaches have been previously proposed that de-bias a pre-trained model. While the majority of current state-of-the-art debiasing methods focus on changes to the training regime, in this paper, we propose data intervention strategies as a powerful yet simple technique to reduce gender bias in pre-trained models. Specifically, we empirically show that by fine-tuning a pre-trained model on only 10 de-biased (intervened) training examples, the tendency to favor any gender is significantly reduced. Since our proposed method only needs a few training examples, our few-shot debiasing approach is highly feasible and practical. Through extensive experimentation, we show that our debiasing technique performs better than competitive state-of-the-art baselines with minimal loss in language modeling ability.

\end{abstract}

\section{Introduction}
Recently, there has been a surge of interest in pre-trained large language models (LLM) in natural language processing (NLP). It has been shown that the pre-training + finetuning of a model drastically improves its performance on downstream tasks as the knowledge captured by the pre-training on a large corpus is transferred to the downstream application when finetuning the model. However, this also leads to societal biases like gender bias that were implicitly learned by the pre-trained models being transferred to crucial downstream applications like job recommendation engines ~\cite{zhao2019gender, barocas2017problem, kurita2019measuring}. Analyzing and mitigating bias without requiring significant re-training or compute resources is crucial to the widespread adoption of LLMs in downstream applications.

Previous work \cite{nadeem-etal-2021-stereoset}, \cite{nangia-etal-2020-crows}, \cite{cer-etal-2018-universal} has attempted to quantify bias, and others such as \citet{ravfogel-etal-2020-null} and \citet{anlp} have attempted to remove it algorithmically from the models. Closer to our work are data-manipulative techniques such as \citet{zmigrod-etal-2019-counterfactual} and \citet{hall-maudslay-etal-2019-name} that modify the dataset and further fine-tune the model. In this paper, we propose simple data intervention strategies and show that they can mitigate gender bias in pre-trained models with the help of few-shot fine-tuning. Moreover, taking inspiration from \citet{self-debias}, we find that by utilizing a biased pre-trained LLM for mining for most gender-biased samples in a dataset, our methods can mitigate gender bias with very few training samples. Finally, we perform an extensive evaluation of our debiasing technique on two recent bias benchmarks \cite{nadeem-etal-2021-stereoset} and show that our method outperforms three existing state-of-the-art techniques and performs comparably to the other two. Our main contributions are the following:

\begin{itemize}
    \item We propose simple data intervention techniques that can be used to reduce gender bias in a pre-trained LLM with few training examples (few-shot), thus making human-in-the-loop bias mitigation strategies feasible.
    \item We introduce a novel data sampling technique that utilises LLMs to mine for the most biased samples from a dataset and can benefit existing state-of-the-art debiasing methods. When used for debiasing a model, these few samples serve as exemplars and induce large reductions in gender bias.
   
\end{itemize}

\begin{figure}
\centering
\includegraphics[scale=0.43]{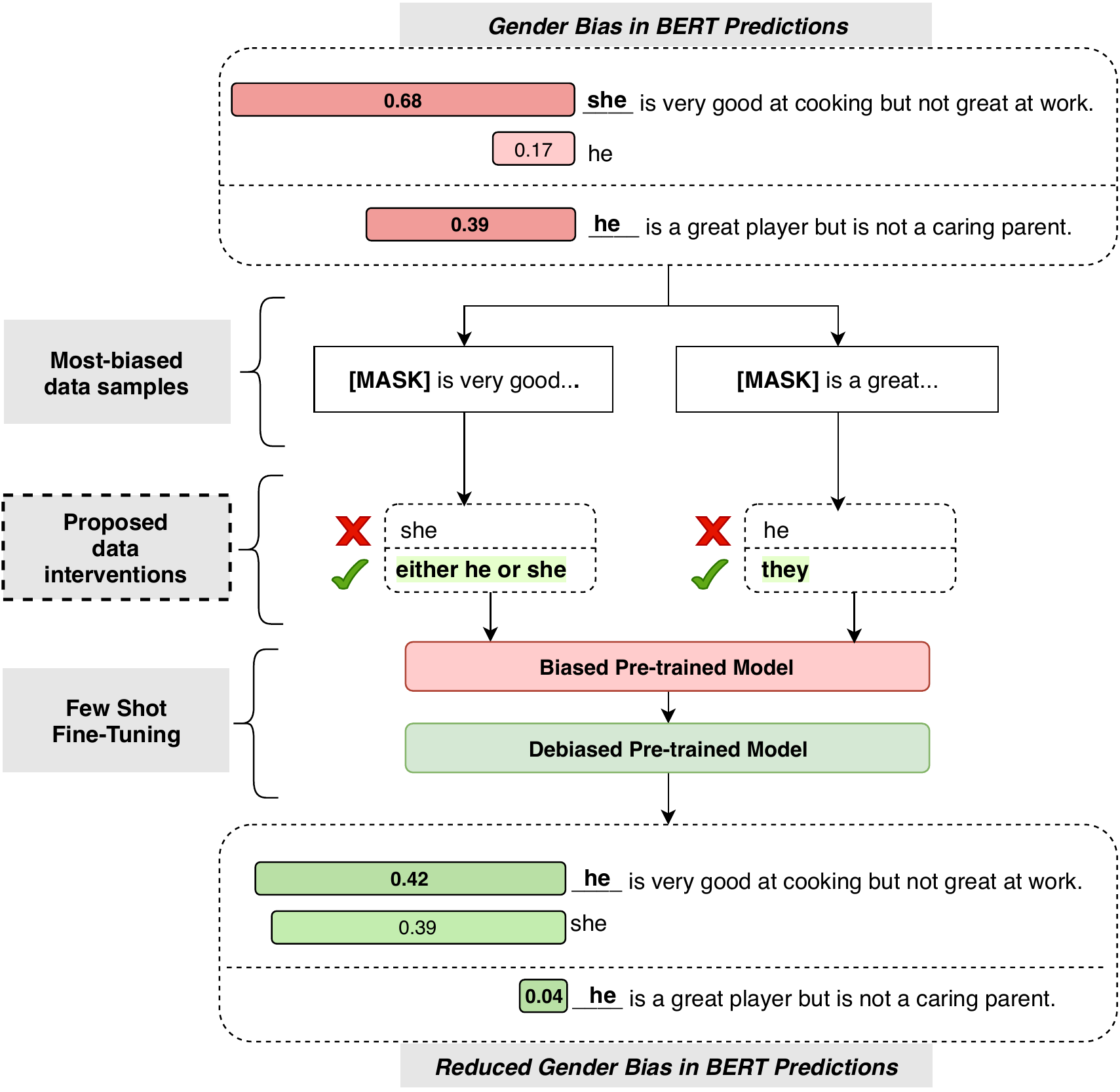}
\caption{Our method can be summarized as a combination of bias discovery and mitigation. First, we use a pre-trained LLM to find 
 the most gender-biased samples. Then, we apply our data intervention techniques and use these modified training samples to fine-tune the model. Experiments show that our method is very effective at reducing gender bias, outperforming three state-of-the-art baselines and being comparable to two other baselines.}

\label{fig:bias1}
\end{figure}

\section{Related Work}

In recent years, there has been growing concern about the bias/stereotypical discriminatory behavior by NLP models, particularly concerning gender. Several studies have investigated the presence of gender bias in various NLP tasks and proposed methods for mitigating it.

One line of research has focused on analyzing the extent of gender bias in pre-trained language models such as BERT and GPT-2. These studies have found that these models exhibit a significant amount of gender bias in their word embeddings for BERT~\cite{jentzsch-turan-2022-gender} and for GPT-2~\cite{https://doi.org/10.48550/arxiv.2102.04130} and are prone to making stereotypical gender-based predictions (e.g., assuming that a doctor is male and a nurse is female). A standard evaluation metric used in this line of research is Stereotype metrics such as StereoSet \cite{nadeem-etal-2021-stereoset}, which evaluates the model's ability to predict gender stereotypes and CrowS pairs \cite{DBLP:journals/corr/abs-2010-00133} which measure whether a model generally prefers more stereotypical sentences. A similar line of work is gender bias tests proposed in BIG-bench \cite{srivastava2022beyond}. The tests assess the language model's gender biases, stereotypes, and ability to infer gender information. It evaluates gender bias and stereotype between male and female, and gender minority bias and stereotype between majority and minority. It also examines the model's language modeling performance, which can be affected during de-biasing.

Another line of research has proposed methods for debiasing these models. These methods can be broadly categorized into two groups: \textbf{data-based} and \textbf{algorithm-based}. Data-based methods aim to reduce bias by removing or altering biased words from the training set. In contrast, algorithm-based methods aim to modify the model's architecture or training procedure to reduce bias. One popular data-based method is "uncertainty sampling" \cite{us}, where the model is trained on the instances that it is most uncertain about, which can help to reduce bias by forcing the model to learn from a diverse set of examples. A popular algorithm-based method is "Adversarial Debiasing" proposed by \citet{DBLP:journals/corr/abs-1801-07593}, which fine-tunes the model using an adversarial loss to make it less sensitive to sensitive attributes such as gender. OSCar proposed by \citet{dev-etal-2021-oscar}, is another algorithm based method that utilizes the idea of disentangling "problematic concepts" like occupation and gender relationship instead of removing them altogether. MABEL~\cite{he2022mabel} has both algorithm and data-based components, as it first augments the training data by swapping gender words and then applies a contrastive learning objective and alignment via entailment pairs. Their data augmentation strategy is similar in spirit to the data intervention techniques we propose, however our analysis does not require training auxiliary models and uses significantly lesser data.

Data-based methods include the "Equalization" technique proposed by \citet{DBLP:journals/corr/BolukbasiCZSK16a}, which aims to equalize the representation of gender-specific words in the embedding space, the "Counterfactual Data Augmentation" (CDA) method proposed by \citet{zimmermann-hoffmann-2022-absinth}, which generates counterfactual examples to improve the model's robustness to bias, and "Name-Based Counterfactual Data Substitution" proposed by \citet{hall-maudslay-etal-2019-name} which reduces gender bias by replacing gender-informative names in the dataset with gender-neutral names. Our proposed method is also a data-based method, which aims to effectively reduce gender bias by taking inspiration from different techniques such as uncertainty sampling and name-based counterfactual data substitution \cite{hall-maudslay-etal-2019-name}.
\section{Probing Bias in Large Language Models}

 Pre-trained LLMs are biased towards different genders, as seen in a simple mask-fill experiment using BERT. (Here, and in the rest of the paper, we assume a binary treatment of gender for simplicity.) The task is then to mask out the gender-related nouns and pronouns (such as he, she, her, woman, etc.) and get BERT to predict the masked words for the affected sequences in the dataset. Here, we consider a fixed list of gender-specific words curated from previous work \cite{wordlist1,zmigrod-etal-2019-counterfactual} and neutral words list\footnote{\url{https://github.com/joelparkerhenderson/inclusive-language}}. We finally compute the "total confidence difference" as the sum of differences in the model's prediction confidence for each gender-word pair (such as confidence of predicting he $-$ she, man $-$ woman, etc.). Formally, we define total confidence difference as $|\sum\nolimits _{i=0}^{N} (f(x_{female}^{(i)} )-f(x_{male}^{(i)} ))|$ where $f(x)$ represent the confidence of model's prediction, $N$ is the total number of tokens in the dataset and $x$ is the tokenized gender word. The higher this number, the more biased the model is concluded to be. We compute the metric at token level and ensure that each of the gender word gets tokenized into exactly one token by initially extending the tokenizer with our gender word list. The top 3 biased gender-word pairs in StereoSet are shown in Table \ref{table:conf-diff}.
Intuitively, our technique for gauging bias in LLMs is sensitive to the fixed word list used to represent the sensitive attributes (here, gender). In Table~\ref{table:word-freq}, we show the number of words covered by the word list used for both WikiText-2 and StereoSet datasets. 

\begin{table}[]
\centering
\resizebox{4.5cm}{!}{
\begin{tabular}{@{}ccc@{}}
\toprule
\multirow{2}{*}{\begin{tabular}[c]{@{}c@{}}Gender-Word \\ Pairs\end{tabular}} & \multicolumn{2}{c}{\begin{tabular}[c]{@{}c@{}}Mean\\ Confidence \\ Difference\end{tabular}} \\ \cmidrule(l){2-3} 
                                                                              & Mean                                       & Std. Dev.                                      \\ \midrule
he, she                                                                       & 0.317                                      & 0.288                                          \\ \midrule
Will, May                                                                     & 0.316                                      & 0.225                                          \\ \midrule
boy, girl                                                                     & 0.219                                      & 0.218                                          \\ \bottomrule
\end{tabular}}
\caption{Confidence difference for the Top 3 gender-word pairs in StereoSet}
\label{table:conf-diff}

\end{table}

\begin{table}
\centering
\resizebox{6cm}{!}{
\begin{tabular}{lcc}
\hline
Dataset                     & Samples & \begin{tabular}[c]{@{}c@{}}Affected Words\\ (mean)\end{tabular} \\ \hline
\multirow{3}{*}{WikiText-2} & 10      & 191                                                             \\ \cline{2-3} 
                            & 50      & 627                                                             \\ \cline{2-3} 
                            & 100     & 1028                                                            \\ \hline
\multirow{3}{*}{StereoSet}  & 10      & 55                                                              \\ \cline{2-3} 
                            & 50      & 227                                                             \\ \cline{2-3} 
                            & 100     & 463                                                             \\ \hline
\end{tabular}}
\caption{Number of words (mean ) covered by the word list vs dataset and number of sequences sampled from each dataset}
\label{table:word-freq}

\end{table}
\vspace{2.5em}

\vspace{-2em}
\section{Data Interventions}

In order to reduce gender bias in pre-trained models, we carefully select diverse and hard-biased examples and then replace gender words with more neutral or equality-focused phrases. This is achieved by using a wordlist to find gender terms in sentences and then segregating words as name and non-name words.

We call our initial approach \texttt{naive-masking} as it does not require a word list for mapping gender words to gender-neutral words. Instead, it replaces all gender words with the fixed word "person." In our next approach, \texttt{neutral-masking}, we swap words in a slightly more semantically accurate manner. In this, we use a word-pair list that goes from gender words to gender-neutral words. With both approaches, we intend to introduce new words in a model's vocabulary to make it more likely to choose a more neutral word in gender-biased sentences. 

In our final approach, we exploit the existing vocabulary of the model and try to balance the confidence of prediction on opposite-gender words by using phrases instead. Thus, we call our final approach \texttt{random-phrase-masking} as we instead substitute words with phrases that reflect the equality of gender. 
This approach not only reduces gender bias but also preserves the original meaning of the sentence in most cases. In our approach, we chose the phrases and order of gender words at random with equal probability.

\begin{table}[h]
\resizebox{7.8cm}{!}{
\centering
\begin{tabular}{@{}ccc@{}}
\toprule
Intervention                     & Input word & Converted word \\ \midrule
\multirow{3}{*}{\texttt{naive-masking}} & he & person \\ 
                            & she & person \\                                                                    
                            & boy & person \\ \midrule 
\multirow{3}{*}{\texttt{neutral-masking}}  & he & they \\
                                  & her & their \\
                                  & schoolgirl & schoolkid \\ \midrule
\multirow{3}{*}{\texttt{random-phrase-masking}} & he & he or she \\ 
                                        & she & she and he \\ 
                                        & boy & either girl or boy \\ \bottomrule 
\end{tabular}}
\label{word-freq}
 \setlength{\belowcaptionskip}{-15pt}
\caption{ Example conversions for three methods. In Random Phrase Masking, the phrase is being chosen and it's order was random.}
\end{table}


Additionally, we hypothesize that the choice of the dataset for fine-tuning is also essential. We choose two datasets: the WikiText-2 \cite{merity2017pointer} dataset, which has implicit gender bias since its sources from Wikipedia articles, and the StereoSet dataset \cite{nadeem-etal-2021-stereoset},  which has explicit/more gender bias as it has been designed to evaluate gender bias. WikiText-2\footnote{An English language dataset (Creative Commons Attribution-ShareAlike License).} has 600 train articles and roughly 2M tokens while StereoSet\footnote{An English language dataset available at \hyperlink {https://github.com/McGill-NLP/bias-bench}{bias-bench} (Creative Commons Attribution-ShareAlike 4.0 International Public License)} (dev) has 2123 samples out of which we only consider 800 samples which are not unrelated. Naturally, our data intervention method should work better on a dataset with training examples with gender bias while being devoid of meaningful gender associations like "She needs a gynecologist," where the gender of the person is important. By testing our method on both datasets, we can understand the sensitivity of our approach to the quality of training samples used.

\section{Bias Evaluation Metrics}

We focus on evaluating the bias of a model while also measuring its language modeling capability. The ideal model would not just be one with the least bias but also one which does not compromise its language modeling performance. The dual estimation of bias and performance of a model was proposed in the StereoSet benchmark \cite{nadeem-etal-2021-stereoset}, with the Language Modeling Score (LMS) measuring the percentage of times a meaningful token is predicted for the mask as opposed to a meaningless token, the Stereotype Score (SS) measuring the percentage of times the model predicted a stereotypical word as compared to an anti-stereotypical word, and an idealized CAT score (ICAT) combining the LMS and SS score into a single metric. An ideal model has an ICAT score of 100, while the worst biased model has an ICAT score of 0. We additionally evaluate the CrowS-Pairs benchmark \cite{nangia-etal-2020-crows}, which captures data with greater diversity in both the stereotypes expressed and the structure of sentences (50 is ideal). However, we note that the Crow-S benchmark is much more limited compared to StereoSet \cite{nadeem-etal-2021-stereoset} in terms of both the volume and variety of linguistic phenomenon relating to gender bias it covers.

\vspace{-0.1em}
\section{Experiments}
We compare our proposed interventions with five baselines, 4 of which are state-of-the-art methods and the original pre-trained model. Our first baseline is the application of dropouts to neural networks, \textbf{Dropout} proposed by \cite{dropout}. Next, we consider an algorithmic de-biasing technique \textbf{INLP} technique proposed by \cite{ravfogel-etal-2020-null}. Then, we consider a sentence embedding de-biasing approach \textbf{SentenceDebias} \cite{liang-etal-2020-towards}. Finally, we consider a data-based approach \textbf{CDA} \cite{zmigrod-etal-2019-counterfactual} that is closest to our work. For a fairer comparison, we run the baselines with the same size (100) of the training set as our method. For all of our experiments, we consider the “bert-base-uncased” pre-trained model available from HuggingFace. For fine-tuning our model, we select a varying number of most-biased training samples (10, 50, and 100) from the WikiText-2 and StereoSet (we only use the dev set) datasets, as discussed in section 4. We also compare this to a random selection of data points as an ablation study. On the selected dataset, we apply our interventions and obtain the modified dataset, which is then used to fine-tune our pre-trained model using masked language modeling (MLM) loss. The key point is that we only fine-tune the model on the gender words conditioned on the remaining text, significantly reducing the fine-tuning time. We perform ablations on various types of interventions as discussed in Table \ref{table:phrases}. The model is trained for 30 epochs, with a learning rate of 0.001 and AdamW optimizer. We ran all of our experiments on NVIDIA Tesla T4 GPU on Google Colab for roughly 48 hours. For all experiments, we report the numbers as the mean and standard deviations (\ref{table:all_runs}) of 3 different runs. Our experiment code can be found here.\footnote{\url{https://github.com/himansh005/data_debias}}

\begin{figure*}
\centering
\includegraphics[scale=0.825]{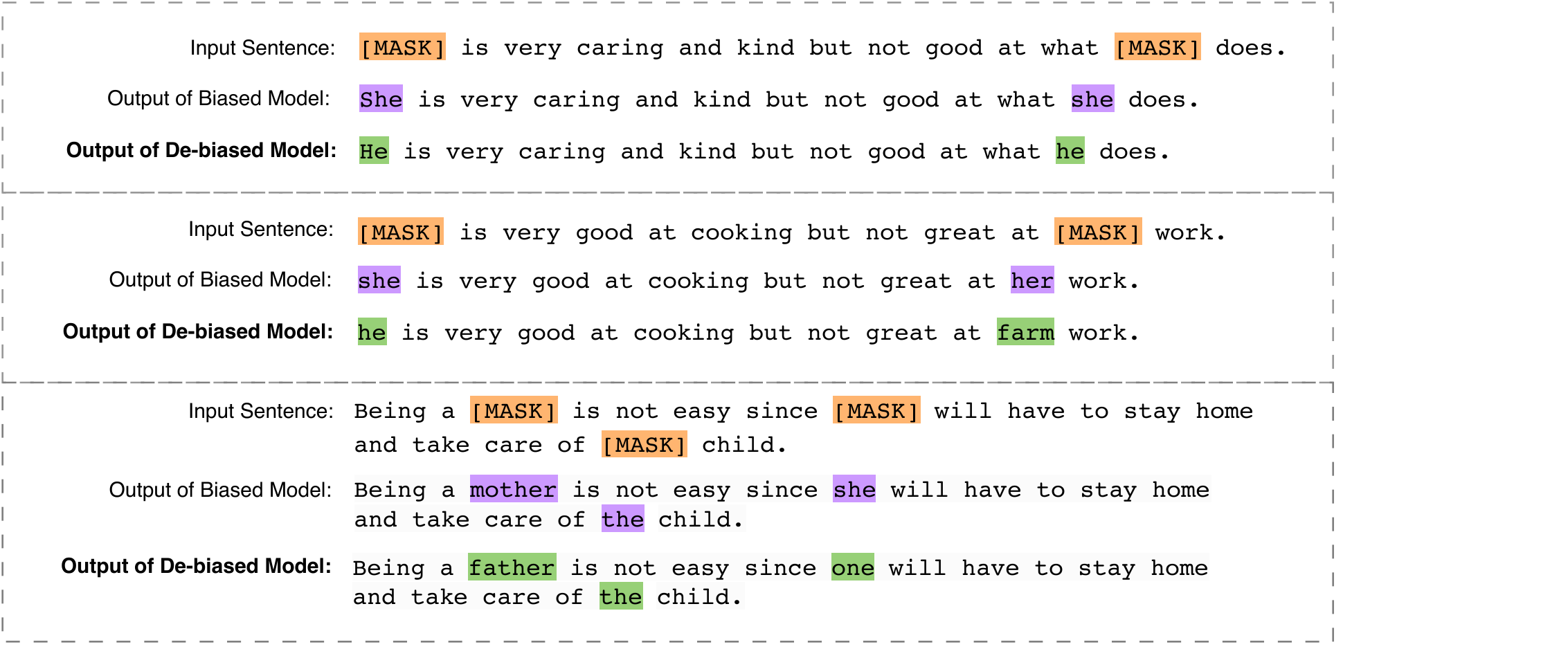}
\caption{Qualitative analysis of our approach on fill-mask task shows that our intervention techniques are able to modify stereotypical sentences. In the this example, we prompted a pre-trained bert-base-uncased model and the same pre-trained model debiased using \texttt{random-phrase-masking} with stereotypical sentences and found that the our method is successfully able to reduced biased substitutions.}

\label{fig:quality}
\end{figure*}

\vspace{2em}
\begin{table}[h]

\centering

\resizebox{7.8cm}{!}{
\setlength\tabcolsep{1.5pt}
\begin{tabular}{@{}
>{\columncolor[HTML]{FFFFFF}}c 
>{\columncolor[HTML]{FFFFFF}}l 
>{\columncolor[HTML]{FFFFFF}}c 
>{\columncolor[HTML]{FFFFFF}}c 
>{\columncolor[HTML]{FFFFFF}}c 
>{\columncolor[HTML]{FFFFFF}}c @{}}

\toprule
\cellcolor[HTML]{FFFFFF} &
  \multicolumn{1}{c}{\cellcolor[HTML]{FFFFFF}} &
  \multicolumn{3}{c}{\cellcolor[HTML]{FFFFFF}StereoSet Scrores} &
  \cellcolor[HTML]{FFFFFF} \\ \cmidrule(lr){3-5}
\multirow{-2}{*}{\cellcolor[HTML]{FFFFFF}Type} &
  \multicolumn{1}{c}{\multirow{-2}{*}{\cellcolor[HTML]{FFFFFF}Method}} &
  SS ($\downarrow$)	 &
  LMS ($\uparrow$)	&
  ICAT ($\uparrow$) &
  \multirow{-2}{*}{\cellcolor[HTML]{FFFFFF}\begin{tabular}[c]{@{}c@{}}Crow-S \\ Scores ($\downarrow$)\end{tabular}} \\ \midrule
\cellcolor[HTML]{FFFFFF}                           & None                       & 60.279          & 84.172          & 70.300          & 57.250          \\ \cmidrule(l){2-6} 
\cellcolor[HTML]{FFFFFF}                           & CDA                       & 60.022          & 83.466          & 70.892          & 56.107          \\ \cmidrule(l){2-6} 
\cellcolor[HTML]{FFFFFF}                           & Dropout                   & 60.529          & 83.811          & 70.171          & 55.977          \\ \cmidrule(l){2-6} 
\cellcolor[HTML]{FFFFFF}                           & SentenceDebias             & 59.221          & \textbf{84.166} & 71.308          & \textbf{53.817}         \\ \cmidrule(l){2-6} 

\multirow{-5}{*}{\cellcolor[HTML]{FFFFFF}\rotatebox[origin=c]{90}{Baselines}} & INLP            &  58.205          & 83.391  & 70.966          & 55.727  \\ \midrule

\cellcolor[HTML]{FFFFFF}                           & \texttt{random-phrase-masking} (10) & 59.442          & 80.312          & 70.406          & 54.580          \\ \cmidrule(l){2-6} 
\cellcolor[HTML]{FFFFFF}                           & \texttt{random-phrase-masking}      & \textbf{58.037} & 78.676          & 69.949          & 54.457 \\ \cmidrule(l){2-6} 
\cellcolor[HTML]{FFFFFF}                           & \texttt{neutral-masking} (10)       & 60.341          & 83.956          & 72.548          & 55.535          \\ \cmidrule(l){2-6} 
\multirow{-4}{*}{\cellcolor[HTML]{FFFFFF}\rotatebox[origin=c]{90}{Ours}}     & \texttt{neutral-masking}            & 60.814          & 83.587 & \textbf{72.213} & 56.490          \\ \bottomrule
\end{tabular}}
 \setlength{\belowcaptionskip}{-15pt}
\caption{StereoSet and Crow-S benchmark results on WikiText-2 dataset. A lower SS and Crow-S score means lesser gender bias while higher ICAT and LMS denote better language modelling ability. The number in the parentheses denotes number of training samples and ones without it use 100.}
\label{table:main-results}
\end{table}
\vspace{-1.5em}
\section{Results}
Table \ref{table:main-results} shows the StereoSet and Crow-S scores for our baselines and our best-performing interventions on the WikiText-2 Dataset. In the StereoSet benchmark, we observe that \texttt{random-phrase-masking} obtains lower SS than all other baselines. On the Crow-S benchmark, \texttt{random-phrase-masking} does better than thre of the baselines except SentenceDebias which achieves slightly better scores. 
While \texttt{random-phrase-masking} results in lower SS scores than \texttt{neutral-masking}, it also obtained very low LMS scores. We attribute this performance degradation to the blunt substitution of phrases that our method uses, which might lead to odd-sounding sentences. In the Crow-S benchmarks, we see similar behavior and find that \texttt{random-phrase-masking} does better than \texttt{neutral-masking}. Since we believe that our method is sensitive to the choice of the dataset, we also present results on the StereoSet (dev) dataset \ref{table:all_runs}. In Figure \ref{fig:quality}, we perform a qualitative analysis of our proposed approach and find that \texttt{random-phrase-masking} is able to flip the predictions on fill-mask tasks for stereotypical sentences.

\section{Conclusion}
In this paper, we show that simple data interventions on limited training data effectively reduce gender bias in LLMs. We also show that a biased pre-trained LLM can be used to mine the most effective de-biasing training examples. Evaluation of our methods on state-of-the-art bias benchmarks empirically suggests that our methods effectively reduce gender bias. Given that our methods can work in a few-shot manner and do not require any auxiliary model training, we hope that our work benefits further research in the domain of human-in-the-loop bias mitigation techniques by making the creation of bias mitigation datasets feasible.

\section{Limitations}

Our proposed method has the following main limitations which we believe are important directions for future work to address:
\begin{enumerate}
    \item \textbf{Gender dependency:} Our approach does not account for sentences that only make sense for a single gender. For example, sentences like "She needs to see a gynecologist" would not be captured by our method. This is a common problem encountered by most debiasing algorithms as it is difficult to distinguish these.
    \item \textbf{Finite wordlist:} The wordlist does not contain all gender-based words as the language continues to evolve. We believe that future works could employ better approaches that can automatically mine gender words relevant to a dataset. 
    \item \textbf{Blunt substitution:} The phrase substitution method is an improvement over direct word substitution, but there are still plenty of instances where the new sentence might be semantically incorrect. This does not have any major implication on inference as we are only doing few-shot learning, but it should not be extended to the entire dataset. 
    \item \textbf{Binary gender:} The method only focuses on the male and female gender. It does not consider non-binary or gender-neutral pronouns such as "ze/hir." This can be solved by using an updated wordlist, but the authors could not come across one at the time of writing.

    \item \textbf{Downstream analyses:} While our work proposes methods that show reduced gender bias as per a set of metrics, the work in no way claims to reduce gender bias in general, especially on downstream tasks.  However, we strongly believe that this technique holds potential to reduce gender bias on downstream tasks as well since we adopt a regular fine-tuning approach and focus mainly on better data interventions. Moreover, recent research has shown that fine-tuning-based debiasing approaches do not damage a model’s internal representations to a critical extent \cite{meade_empirical_2022}.

\end{enumerate}

Overall, these limitations suggest that our approach may not be suitable for use in contexts where gender-specific or non-binary language is prevalent, and the underlying wordlist should be frequently updated.

\section{Ethics Statement}

This study was conducted in accordance with ethical principles and guidelines. The study was designed to provide beneficial knowledge and not harm any group or individual. We recognize that the wordlist we use might not represent all contexts of gender bias and that our debiasing method does not cover all contexts of occurrences of gender bias. However, we made sure to consider the ethical implications of our methodologies and the results of our analysis. The authors have tried to ensure the method does not amplify any other inherent bias but also acknowledge that our approach may have limitations. We take responsibility for any ethical concerns that may arise as a result of our research.

\vspace{-0mm}
\section*{Acknowledgments}

This material is based upon work partially supported by the National Science Foundation (Awards \#1722822 and \#1750439) and National Institutes of Health (Awards \#R01MH125740, \#R01MH096951, and \#U01MH116925). PPL is partially supported by a Facebook PhD Fellowship and a Carnegie Mellon University's Center for Machine Learning and Health Fellowship.
Any opinions, findings, conclusions, or recommendations expressed in this material are those of the author(s) and do not necessarily reflect the views of the NSF, NIH, Facebook, or CMLH, and no official endorsement should be inferred.
Additionally, we express our appreciation to the anonymous reviewers for their insightful suggestions, which greatly improved our work. Furthermore, we would like to acknowledge the contributions of our colleagues, Atishay Jain and Praneetha Vaddamanu, who played a significant role in the development of this research.

\bibliography{anthology,custom}
\bibliographystyle{acl_natbib}

\appendix

\section{Appendix}

\subsection{Dataset Bias Analysis}

To gauge the feasibility of using a wordlist based intervention approach, we first analyze our datasets for occurrences of gender words. As shown in the word cloud \ref{fig:word_cl}, gender pronouns are the most-frequent word in our datasets. Moreover, as per Figure \ref{fig:bias1}, "she," "he," and "her" are the top three most frequently occurring words in our dataset. This suggests that we can definitely detect gender words in our corpus and apply our interventions.

\begin{table*}[]
\centering
\begin{tabular}{@{}cc@{}}
\toprule
Sentences                                                                                                                                                                                                                            & \begin{tabular}[c]{@{}c@{}}Mean\\ Confidence\\ Difference\end{tabular} \\ \midrule
\begin{tabular}[c]{@{}c@{}}She rushed to see what he wanted and said she loved him. \\ She punched him in the face and told him to go away.\end{tabular}                                                                             & 6.85                                                                   \\ \midrule
\begin{tabular}[c]
{@{}c@{}}Jessica is a new mommy. \\ Jessica finds being a mother does not come easy to her. \\ She will no longer work so she can stay home and take care of her child.\end{tabular} & 6.34                                                                   \\ \midrule
\begin{tabular}[c]{@{}c@{}}The little girl missed her mommy.\\ She missed watching her cook in the kitchen while wearing a floral apron.\\ She was never home because she worked long hours in the oil field.\end{tabular}           & 4.70                                                                   \\ \bottomrule
\end{tabular}
\label{table:most-bias-sent}
\caption{Sentences from StereoSet with maximum difference in confidence of prediction between opposite gender words.}
\end{table*}
\begin{figure}[H]
\centering
\includegraphics[scale=0.5]{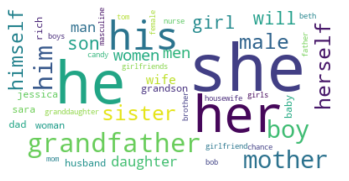}
\caption{Frequency of gender words on the StereoSet dataset.}
\label{fig:word_cl}
\end{figure}

\begin{figure}[H]
\centering
\includegraphics[scale=0.45]{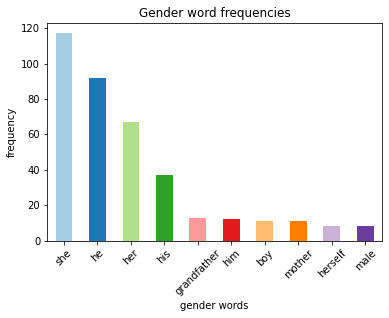}
\caption{Top 10 most frequent gender words on the StereoSet dataset.}
\label{fig:word_cl}
\end{figure}
\subsection{Sensitivity to Choice of Dataset}

To understand the effectiveness of our proposed data-interventions, we study apply our methods to two datasets under varying number of training samples (10, 50 and 100) and selection strategies (most biased first and random) as per Table \ref{table:all_runs}. Our methods obtain better results on StereoSet (dev) dataset. One reason this could happen is due to the fact that StereoSet has explicit gender bias, thus it would be less likely for a sentence like "She needs a gynaecologist" to appear on it. Because our interventions perform blunt substitutions, this sentence might become incorrect due to our method - "Either he or she needs a gynaecologist".

\begin{table*}[]
\centering
\resizebox{16cm}{!}{
\begin{tabular}{@{}cccccccccc@{}}
\toprule
\multirow{2}{*}{Dataset} &
  \multirow{2}{*}{\begin{tabular}[c]{@{}c@{}}Sampling\\ Method\end{tabular}} &
  \multirow{2}{*}{\begin{tabular}[c]{@{}c@{}}Number of\\ Samples\end{tabular}} &
  \multicolumn{3}{c}{Crow-S Pair Score} &
  \multicolumn{3}{c}{StereoSet Scores} &
  \multirow{2}{*}{Perplexity} \\ \cmidrule(lr){4-9}
 &
   &
   &
  Total &
  Stereotype Score &
  Anti-stereotype Scoe &
  SS (gender) &
  LMS (gender) &
  ICAT (gender) &
   \\ \midrule
\multirow{6}{*}{StereoSet} &
  most-biased &
  10 &
  54.481 (2.583) &
  50.408 (5.295) &
  60.991 (3.854) &
  58.736 (1.215) &
  80.858 (2.988) &
  66.708 (2.584) &
  50.449 (54.983) \\ \cmidrule(l){2-10} 
 &
  random &
  10 &
  55.47 (3.247) &
  50.527 (3.632) &
  63.107 (4.234) &
  58.952 (0.859) &
  80.226 (2.85) &
  65.862 (2.655) &
  86.024 (107.709) \\ \cmidrule(l){2-10} 
 &
  most-biased &
  50 &
  \textbf{52.994 (1.894)} &
  47.567 (4.564) &
  61.428 (5.25) &
  58.498 (1.19) &
  80.255 (2.428) &
  66.595 (2.207) &
  29.599 (28.648) \\ \cmidrule(l){2-10} 
 &
  random &
  50 &
  53.817 (1.011) &
  \textbf{50.107 (2.972)} &
  59.547 (3.925) &
  58.485 (0.758) &
  79.158 (1.992) &
  65.707 (0.886) &
  62.498 (11.593) \\ \cmidrule(l){2-10} 
 &
  most-biased &
  100 &
  53.054 (2.402) &
  49.063 (6.025) &
  59.291 (4.663) &
  58.071 (1.158) &
  81.086 (3.226) &
  67.972 (2.671) &
  \textbf{19.079 (14.095)} \\ \cmidrule(l){2-10} 
 &
  random &
  100 &
  53.563 (1.801) &
  48.113 (6.499) &
  62.137 (5.405) &
  \textbf{57.719 (1.94)} &
  79.038 (1.406) &
  66.805 (2.074) &
  34.826 (12.109) \\ \midrule
\multirow{6}{*}{WikiText-2} &
  most-biased &
  10 &
  55.6 (3.06) &
  54.668 (5.606) &
  57.118 (1.671) &
  59.344 (0.742) &
  84.624 (2.134) &
  \textbf{68.811 (2.176)} &
  87.06 (80.998) \\ \cmidrule(l){2-10} 
 &
  random &
  10 &
  56.617 (1.344) &
  57.983 (1.305) &
  \textbf{54.693 (2.019)} &
  60.616 (0.72) &
  \textbf{85.076 (0.896)} &
  67.021 (1.895) &
  59.901 (102.019) \\ \cmidrule(l){2-10} 
 &
  most-biased &
  50 &
  54.276 (1.513) &
  53.394 (3.847) &
  55.834 (2.652) &
  59.238 (1.068) &
  83.348 (3.003) &
  67.902 (0.977) &
  212.365 (155.526) \\ \cmidrule(l){2-10} 
 &
  random &
  50 &
  54.2 (2.383) &
  51.783 (5.272) &
  57.93 (2.969) &
  59.611 (1.155) &
  83.456 (1.9) &
  67.386 (0.74) &
  116.872 (100.401) \\ \cmidrule(l){2-10} 
 &
  most-biased &
  100 &
  55.473 (1.42) &
  54.827 (4.255) &
  56.637 (4.329) &
  59.426 (1.719) &
  83.442 (3.185) &
  67.629 (1.178) &
  220.957 (207.243) \\ \cmidrule(l){2-10} 
 &
  random &
  100 &
  54.457 (1.444) &
  51.363 (4.283) &
  59.223 (4.451) &
  59.545 (0.387) &
  81.953 (1.442) &
  66.3 (0.597) &
  326.017 (181.822) \\ \bottomrule
\end{tabular}}
\caption{StereoSet and Crow-S scores for  \texttt{random-phrase-masking} method on two datasets, 3 sample sizes and 2 selection methods. We report mean (standard deviation) across 3 different runs. Selecting most-biased samples and using the StereoSet dataset for fine-tuning gives best results. }
\label{table:all_runs}

\end{table*}

\begin{table*}[]
\centering

\resizebox{16cm}{!}{
\begin{tabular}{@{}ccccccccc@{}}
\toprule
\multirow{2}{*}{\begin{tabular}[c]{@{}c@{}}Name Word\\ Mask Method\end{tabular}} &
  \multirow{2}{*}{\begin{tabular}[c]{@{}c@{}}Non-Name Word\\ Mask Method\end{tabular}} &
  \multicolumn{3}{c}{Crow-S Pairs} &
  \multicolumn{3}{c}{StereoSet} &
  \multirow{2}{*}{Perplexity} \\ \cmidrule(lr){3-8}
 &
   &
  Total &
  Stereotype Score &
  Anti-Stereotype Score &
  SS &
  LMS &
  ICAT &
   \\ \midrule
\multicolumn{2}{c}{\begin{tabular}[c]{@{}c@{}}\texttt{female-first-}\\ \texttt{random-phrase-masking}\end{tabular}} &
  53.18 (3.106) &
  49.06 (6.627) &
  59.55 (3.678) &
  58.283 (0.4) &
  79.059 (0.436) &
  65.96 (0.27) &
  26.3 (9.545) \\ \midrule
\multicolumn{2}{c}{naive-masking} &
  \textbf{50.637 (0.585)} &
  43.607 (1.449) &
  61.49 (1.486) &
  59.521 (0.458) &
  83.325 (0.62) &
  67.456 (0.414) &
  \textbf{1.0 (0.0)} \\ \midrule
naive-masking &
  \texttt{random-phrase-masking} &
  52.673 (1.374) &
  49.057 (5.998) &
  58.253 (5.906) &
  58.05 (0.851) &
  78.218 (0.633) &
  65.618 (0.937) &
  30.045 (8.019) \\ \midrule
\texttt{neutral-masking} &
  \begin{tabular}[c]{@{}c@{}}\texttt{female-first-}\\ \texttt{random-phrase-masking}\end{tabular} &
  53.44 (0.0) &
  53.46 (0.891) &
  \textbf{53.4 (1.372)} &
  58.246 (0.285) &
  \textbf{87.182 (0.391)} &
  \textbf{72.806 (0.823)} &
  11.39 (6.649) \\ \midrule
\multicolumn{2}{c}{\texttt{random-phrase-masking}} &
  54.195 (1.619) &
  48.43 (0.891) &
  63.11 (2.744) &
  57.316 (0.164) &
  78.339 (0.196) &
  66.877 (0.424) &
  54.413 (0.212) \\ \midrule
\multicolumn{2}{c}{\texttt{fixed-phrase-masking-1}} &
  53.307 (1.761) &
  46.837 (8.494) &
  63.43 (8.807) &
  57.688 (1.718) &
  79.554 (0.17) &
  67.32 (2.64) &
  14.484 (1.512) \\ \midrule
\multicolumn{2}{c}{\texttt{fixed-phrase-masking-2}} &
  51.783 (4.965) &
  46.43 (10.381) &
  60.193 (3.503) &
  57.229 (1.739) &
  80.551 (1.251) &
  68.879 (1.882) &
  13.374 (1.174) \\ \midrule
\multicolumn{2}{c}{\texttt{fixed-phrase-masking-3}} &
  52.927 (1.541) &
  48.317 (3.78) &
  60.193 (4.234) &
  \textbf{56.963 (1.373)} &
  79.3 (1.531) &
  68.284 (3.478) &
  15.546 (2.997) \\ \midrule
\multicolumn{2}{c}{\texttt{fixed-phrase-masking-4}} &
  53.567 (4.186) &
  \textbf{50.083 (9.006)} &
  59.223 (3.885) &
  58.13 (1.208) &
  79.834 (0.533) &
  66.86 (2.309) &
  14.51 (1.339) \\ \bottomrule
\end{tabular}}
\caption{StereoSet Gender SS scores on StereoSet (dev) dataset on 100 samples across various interventions techniques. All numbers are reported as mean and standard deviation across 3 runs.}
\label{table:phrases}
\end{table*}

\subsection{Sensitivity to Number of Training Samples and Sampling Strategy}

\begin{figure}[H]
\centering
    \begin{center}
        \scalebox{.7}{\input{figures/histogram.pgf}}
    \end{center}
\caption{StereoSet Gender SS scores on StereoSet (dev) dataset with varying number of training samples. Compared to random selection of samples, selecting most-biased samples help achieve lower SS scores.}
\label{fig:num_train}
\end{figure}

As per Figure \ref{fig:num_train}, When we vary the number of training samples, we observe that the difference in performance is not huge when we transition from 10 to 100 samples, thus suggesting that our method is capable of few-shot fine-tuning. Moreover, sampling the most biased data points helps our methods achieve better performance consistently, as shown in Figure \ref{fig:num_train} and Table \ref{table:all_runs}. Table \ref{table:most-bias-sent} shows some top three most gender biased entries found in the StereoSet dataset.

\subsection{Ablations of interventions}

We study the effects of choosing different ways of replacement for name and non-name words. In addition to our three interventions proposed previously, we also experimented with a couple of others. In \texttt{female-first-random-phrase-masking}, we always keep the female gendered word before a male word. We wanted to understand if the order of gender words encountered by a model renders any effect on the debiasing. In Table \ref{table:phrases}, we see that it does not perform any better than \texttt{random-phrase-masking}. Then, we also try fixing the phrases from \texttt{random-phrase-masking}, thus making it \texttt{fixed-phrase-masking}. We obtain 4 variants of this method corresponding to the following four phrases:

\begin{enumerate}
\setlength{\itemindent}{0.2in}
    \item {both [1] and [2]}
    \item {[1] and [2]}
    \item {[1] or [2]}
    \item {either [1] or [2]}
\end{enumerate}

Here, [1] and [2] are substituted with opposite gender words. As we observe in Table \ref{table:phrases}, \texttt{fixed-phrase-masking-3} obtains the lowest StereoSet Gender SS score out of all our intervention methods. Similarily, \texttt{naive-masking} obtains the lowest Crow-S pair score.

\label{sec:appendix}

\end{document}